\documentclass[letterpaper, 10 pt, conference]{ieeeconf}  

\IEEEoverridecommandlockouts                              

\usepackage{graphics} 
\usepackage{epsfig} 
\usepackage{amsmath} 

\usepackage{multirow}

\usepackage{amsfonts}    
\usepackage{booktabs}
\usepackage{xcolor}

\usepackage{algorithm}   
\usepackage[noend]{algpseudocode} 

\floatname{algorithm}{Procedure}

\usepackage{amssymb}
\usepackage{pifont}
\newcommand{\cmark}{\ding{51}}%
\newcommand{\xmark}{\ding{55}}

\makeatletter
\let\NAT@parse\undefined
\makeatother
\usepackage{hyperref}
\hypersetup{colorlinks,citecolor={blue}}

\title{Towards Multimodal Multitask Scene Understanding Models for \\ Indoor Mobile Agents}

\author{%
  Yao-Hung Hubert Tsai, Hanlin Goh, Ali Farhadi, Jian Zhang\thanks{Apple,
  \texttt{\{yaohung\_tsai,hanlin,afarhadi,jianz\}@apple.com}}
}

\begin{document}

\maketitle
\thispagestyle{empty}
\pagestyle{empty}

\begin{abstract}
    The perception system in personalized mobile agents requires developing indoor scene understanding models, which can understand 3D geometries, capture objectiveness, analyze human behaviors, etc. Nonetheless, this direction has not been well-explored in comparison with models for outdoor environments (e.g., the autonomous driving system that includes pedestrian prediction, car detection, traffic sign recognition, etc.). In this paper, we first discuss the main challenge: insufficient, or even no, labeled data for real-world indoor environments, and other challenges such as fusion between heterogeneous sources of information (e.g., RGB images and Lidar point clouds), modeling relationships between a diverse set of outputs (e.g., 3D object locations, depth estimation, and human poses), and computational efficiency. Then, we describe MMISM (Multi-modality input Multi-task output Indoor Scene understanding Model) to tackle the above challenges. MMISM considers RGB images as well as sparse Lidar points as inputs and 3D object detection, depth completion, human pose estimation, and semantic segmentation as output tasks. We show that MMISM performs on par or even better than single-task models; e.g., we improve the baseline 3D object detection results by 11.7\% on the benchmark ARKitScenes dataset. 
\end{abstract}

\section{Introduction}
\label{sec:intro}

At the core of the perception system in autonomous or intelligent agents is a scene understanding model that acts to process the multi-sensory signals from the environment, generating diversified outputs according to human requests. A well-known example is an autonomous driving system, where the inputs are images from RGB cameras, point clouds from Lidar sensors, and electromagnetic waves from Radar sensors, and the outputs are pedestrian prediction, vehicle tracking, and traffic sign recognition~\cite{caesar2020nuscenes,sun2020scalability}. Although there are recent trends in developing efficient and high-performance scene understanding models, the focus is mostly on outdoor environments~\cite{caesar2020nuscenes,sun2020scalability,geiger2013vision,li2021hdmapnet,liu2022bevfusion}, while the study in indoor environments has not been well-explored. In the following, we identify several challenges for developing scene understanding models for indoor mobile agents. 

The main challenge is insufficient, or even no, labeled data in real-world indoor environments. Specifically, none of the existing indoor scene understanding datasets~\cite{li2021igibson,ramakrishnan2021hm3d,chang2017matterport3d,straub2019replica,baruch2021arkitscenes} provide all data labels for tasks encompassing depth completion, semantic segmentation, 3D object detection, and human pose estimation. To deal with this challenge, in this paper, we start with the ARKitScenes dataset~\cite{baruch2021arkitscenes} (a large-scale indoor dataset with images and Lidar points) that provides sparse depth and 3D object detection labels. We then combine techniques including self-supervised sparse-to-dense depth completion~\cite{ma2019self}, knowledge distillation on pre-trained image segmentation models~\cite{cheng2021masked,ronneberger2015u}, and training on external human pose estimation datasets~\cite{lin2014microsoft,zhou2019objects} to create an indoor scene understanding model. 

In addition to the main challenge, there is also a challenge of how to fuse multi-sensory inputs. Different sensors provide heterogeneous sources of information, such as RGB cameras providing dense 2D color information and Lidar sensors providing sparse 3D location information. A proper fusion between various features brings complementary information together, yet the difficulty lies in how we align features in different spaces~\cite{wang2019pseudo}, with different structures~\cite{tsai2019multimodal}, or with different densities~\cite{qi2018frustum}. In this paper, we consider sparse Lidar points and RGB images from mobile agents as multi-sensory inputs, fusing them for depth completion and 3D object detection tasks.  

Another challenge is modeling the relationship among multi-task outputs. Particularly, tasks are often correlated in scene understanding models. For instance, 3D object detection aims to detect an object's 3D location, which may benefit from depth estimation. Similar to performing fusion 
between multi-sensory inputs, relationship modeling among multi-task outputs has the challenge that different task outputs may lie in different spaces, structures, or densities. We also must be cautious about training multi-task models, as the training can be unstable in comparison to training on single-task models~\cite{ruder2017overview}. In this paper, we find that 1) semantic segmentation outputs can help improve 3D object detection and 2) the multi-task model performs on par or even better than the single-task models.

The last challenge is computational efficiency. Since we aim to design scene understanding models for indoor mobile agents, we must ensure the model is efficient, especially during inference. This demand becomes even more challenging when we consider a multi-task setting. In this paper, we present to share feature backbones across different tasks instead of stacking individual single-task models; otherwise, the computation will be too high.

Putting everything together, we introduce the {\bf M}ulti-modality input {\bf M}ulti-task output {\bf I}ndoor {\bf S}cene understanding {\bf M}odel (MMISM), which to our knowledge is the first scene understanding model for indoor environments that provides a combination of depth completion, semantic segmentation, 3D object detection, and human pose estimation. We provide quantitative and qualitative results across all tasks in the experiments section. We find that MMISM achieves on-par or better performance than single-task models. In particular, MMISM improves the baseline 3D object detection by 11.7\% on the benchmark ARKitScenes dataset~\cite{baruch2021arkitscenes}.

\vspace{-1.5mm}
\section{Related Work}
\label{sec:related}
\vspace{-1.5mm}

Popular outdoor scene understanding datasets, such as Kitti~\cite{geiger2013vision}, NuScenes~\cite{caesar2020nuscenes}, Argo~\cite{wilson2021argoverse}, and Waymo Open~\cite{sun2020scalability} datasets, contain a diverse set of output tasks, advancing self-driving autonomous agents on vehicle detection, pedestrian detection, vehicle tracking, human behavior prediction, etc. Popular indoor scene understanding datasets, such as ScanNet~\cite{dai2017scannet}, Matterport3D~\cite{chang2017matterport3d},  Replica~\cite{straub2019replica}, and ARKitScenes~\cite{baruch2021arkitscenes} datasets, are usually combined with simulated platforms, such as Habitat~\cite{savva2019habitat} and iGibson~\cite{li2021igibson}, for facilitating embodied artificial intelligence, including goal navigation and instruction following. Between existing indoor and outdoor datasets, there are two key differences. First, the types of objects are different. Outdoor datasets contain a lot of moving objects (vehicles and pedestrians), and indoor datasets contain more non-moving objects (furniture, appliance, or office supplies)\footnote{Indoor environments can contain humans. Nonetheless, existing indoor datasets do not focus on providing annotations for humans.}. Objects in indoor environments can be very small, such as a pencil or a brooch (specifically, small relatively at room scale). The difference between object types imposes distinct challenges for indoor or outdoor object detection. Second, the scales of the environment are different. Outdoor environments are assumed to be infinitely large, while indoor environments are limited by the size of rooms or halls. The scale difference makes depth estimation notoriously challenging for indoor scenes and very different from outdoor scenes~\cite{ji2021monoindoor}; especially when we want to obtain accurate depth using less powerful mobile hardware. This paper focuses on studying indoor scene understanding. It is worth noting that we propose to add human pose estimation capabilities for indoor mobile agents by leveraging additional labels in the external MS-COCO dataset~\cite{lin2014microsoft}.

Multi-modal multi-task scene understanding networks~\cite{liang2019multi}, \cite{li2021hdmapnet,liu2022bevfusion,yang2018end,huang2020multi,kendall2018multi} consider (multiview) images and Lidar point clouds as multi-modal inputs, and object detection~\cite{liang2019multi,li2021hdmapnet,liu2022bevfusion}, semantic segmentation~\cite{liang2019multi,li2021hdmapnet,liu2022bevfusion,yang2018end}, trajectory tracking~\cite{li2021hdmapnet,liu2022bevfusion}, and vehicle controls~\cite{yang2018end,huang2020multi} as multi-task outputs. Nonetheless, all of these networks are deployed in outdoor environments. Since indoor and outdoor environments are very different, the models cannot be trivially transferred from outdoor to indoor environments. To deploy a multi-task indoor scene understanding model, a naive approach would be to stack existing indoor single-task models together, such as depth completion~\cite{ji2021monoindoor}, semantic segmentation~\cite{cheng2021masked,he2017mask}, human pose estimation~\cite{martinez2017simple,zhou2019objects}, object detection~\cite{qi2017pointnet,qi2017pointnet++,qi2018frustum,qi2019deep}, etc. However, naively stacking different models together may be too expensive for mobile compute and does not leverage correlation between different tasks. To address these concerns, our work models the relationships between different tasks and presents a feature-backbone-sharing structure.

Our work is also closely related to indoor scene construction engines~\cite{savva2019habitat,li2021igibson,hughes2022hydra}, which render 3D scene graphs according to the visual images, IMU data, depth, object labels, semantic labels, etc. Hence, our model can be integrated with these indoor scene construction engines, potentially with on-device hardware.

\begin{figure*}[t!]
    \centering
    \includegraphics[width=0.9\linewidth]{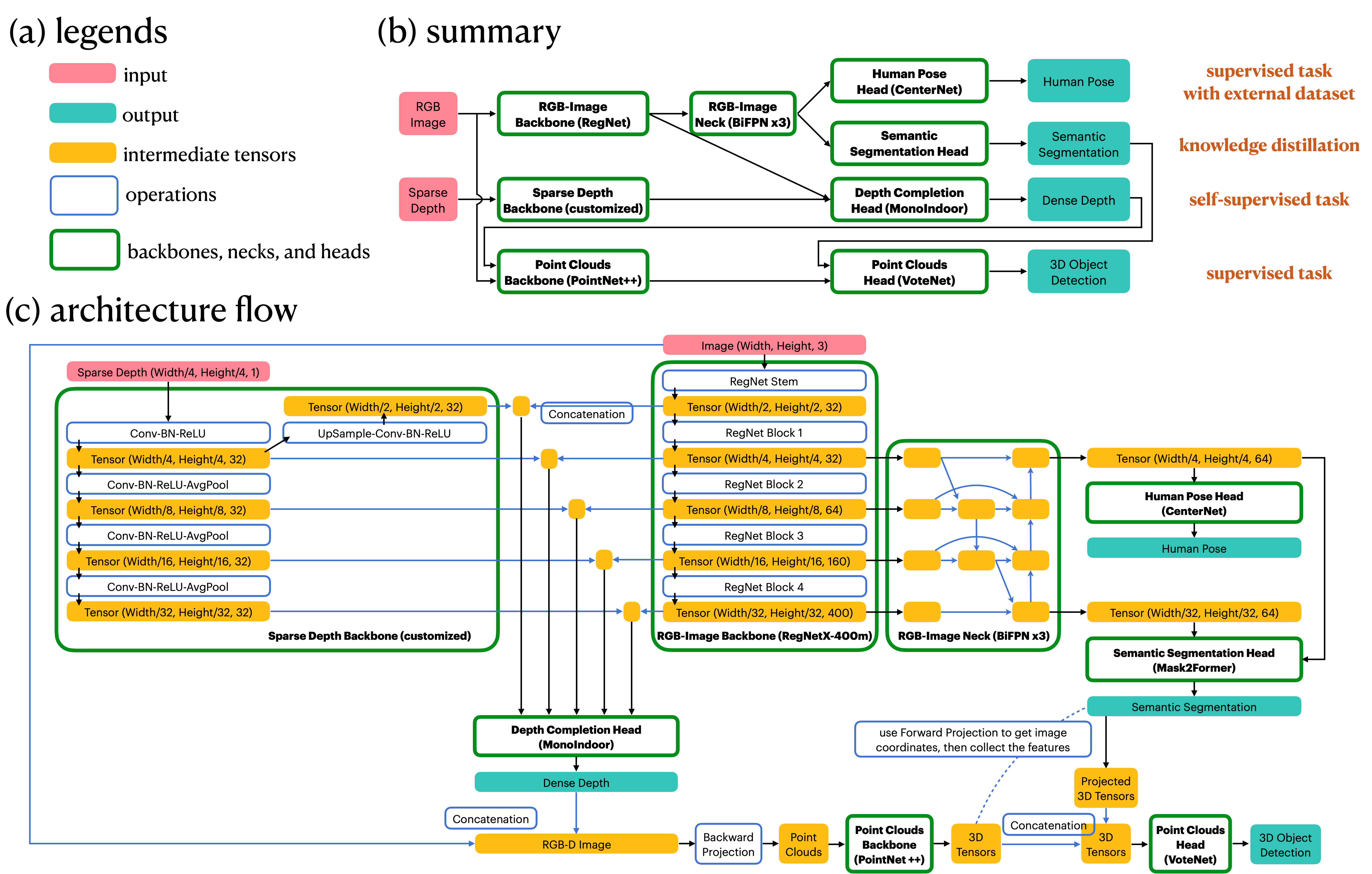}
    \caption{Overview of our {\bf M}ulti-modality input {\bf M}ulti-task output {\bf I}ndoor {\bf S}cene understanding {\bf M}odel (MMISM).}
    \label{fig:illus}
\end{figure*}
\section{Proposed Model}
\label{sec:proposed}

In this paper, we propose {\bf M}ulti-modality input {\bf M}ulti-task output {\bf I}ndoor {\bf S}cene understanding {\bf M}odel (MMISM).
We present a high-level summary of MMISM in Figure~\ref{fig:illus} (b). First, MMISM considers RGB-images and sparse Lidar points as input, and backbones (e.g., the RGB-image and sparse depth backbone) are shared for all tasks. Second, MMISM considers a diverse set of output tasks, where each task is trained differently. Specifically, 1) the human pose estimation task considers supervised training with an external dataset, since the primary dataset contains no humans; 2) the semantic segmentation task is trained using knowledge distillation from a teacher model, since the primary dataset does not contain semantic labels; 3) the dense completion is trained using self-supervision without the need for dense depth labels; and 4) the 3D object detection task considers standard supervised training. In the following, we will describe our data, training, and inference in Section~\ref{subsec:data_and_train}, elaborating on the backbone, task heads, and cross-task fusion in Sections~\ref{subsec:depth},~\ref{subsec:hum_pose},~\ref{subsec:sem_seg}, and~\ref{subsec:object_detec}.

\subsection{Data, Training, and Inference}
\label{subsec:data_and_train}

We select ARKitScenes~\cite{baruch2021arkitscenes} as our primary dataset for three reasons: 1) ARKitScenes is one of the largest released indoor scene understanding datasets; 2) ARKitScenes contains diverse data from rooms in houses across different countries and socioeconomic statuses; and 3) ARKitScenes data is collected using mobile hardware, making it suitable for studying scene understanding models for indoor mobile agents. ARKitScenes provides inputs for MMISM: RGB images (256$\times$192 resolution) and sparse Lidar points (64$\times$48 Lidar points evenly sub-sampled from the depth image using ARKit SDK\footnote{\url{https://developer.apple.com/documentation/arkit}}).

Since ARKitScenes comes with depth information and 3D object labels, we can perform and evaluate depth completion and 3D object detection tasks. Nonetheless, it does not directly support semantic segmentation and human pose estimation tasks. To provide semantic segmentation cability in MMISM, we apply the pre-trained semantic segmentation model, Mask2Former~\cite{cheng2021masked} trained on ADE20K dataset~\cite{zhou2019semantic}, on ARKitScenes images; performing knowledge distillation~\cite{hinton2015distilling} from Mask2Former's output to MMISM's output. To add human pose estimation, we rely on the external MS-COCO human pose dataset~\cite{lin2014microsoft}. 

\underline{\it Training Procedure.} Now we discuss the training procedure for MMISM, and readers can refer to Figure~\ref{fig:illus} (b). 
\begin{enumerate}
    \item Perform inference of the Mask2Former~\cite{cheng2021masked} model to obtain teacher labels for the semantic segmentation task.
    \item Perform a forward-pass for the rgb images and the sparse lidar points to the non-pointcloud backbones and heads, obtaining predictions for the pose estimation, semantic segmentation, and depth completion tasks.
    \item Combine RGB images and the predicted dense depth, performing a forward-pass of it to the point cloud backbone, fusing the point cloud features with predicted semantic segmentation labels. Then, perform a forward-pass for the fused features, obtaining predictions for the 3D object detection. 
    \item Perform back propagation between all predictions and (teacher) labels. 
\end{enumerate}
Our training is performed on 8 Nvidia V100 nvlink gpus with a btach size of $96$ for $100$ epochs. We consider the following loss:
\begin{equation*}
\begin{split}
\mathcal{L}_{\rm Total} = &0.05\cdot \mathcal{L}_{\rm Depth\,\,Completion} + \mathcal{L}_{\rm Semantic\,\, Segmentation}   \\
& +  \mathcal{L}_{\rm Object\,\,Detection} + \mathcal{L}_{\rm Human\,\,Pose\,\,Estimation}
\end{split}
\end{equation*}
We choose the Adam optimizer with a learning rate of $1e-3$ with the exception of $1e-4$ for the depth completion head. We consider the same training and test split released in ARKitScenes. We only use the training split in MS-COCO. 

\underline{\it Inference.} We test the inference on two different setups. On a Nvidia V100 nvlink gpu, the inference for MMISM takes $63$ ms, pose estimation alone takes $14$ ms, semantic segmentation alone takes $14$ ms, depth completion alone takes $12$ ms, and 3D object detection alone takes $48$ ms. On a Nvidia RTX-3080 gpu, the inference for MMISM takes $47$ ms, pose estimation alone takes $5$ ms, semantic segmentation alone takes $5$ ms depth completion alone takes $4$ ms, and 3D object detection alone takes $41$ ms. We can see that the 3D object detection task is the current bottleneck for inference efficiency. Note that our future work focuses on not only improving the efficiency of the 3D object detection task but also speeding up neural network inference using TensorRT or onnx.

\subsection{Depth Completion}
\label{subsec:depth}
 
We consider a self-supervised depth completion setup~\cite{ma2019self} that takes sparse Lidar points (64$\times$48 points) and monocular images (256$\times$192 resolution) as inputs, and outputs dense depth images (depth values for 256$\times$192 resolution). Note that the self-supervised setup avoids the need for dense depth as supervision signals. We show in Figure~\ref{fig:illus} that our depth completion network comprises a RGB-image backbone, a sparse depth backbone, a RGB-image neck, and a depth completion head. We consider the RegNetX-400m~\cite{radosavovic2020designing} as the RGB-image backbone (denoted as $\mathsf{backbone}_{\rm RGB}$), and a customized sparse depth backbone (illustrated in Figure~\ref{fig:illus} (c) and denoted as $\mathsf{backbone}_{\rm sp\_D}$). We consider the depth completion head (denoted as $\mathsf{head}_D$) in the MonoIndoor~\cite{ji2021monoindoor} approach, which contains a depth estimation network (denoted as $\mathsf{head}_D^{depth}$) and a pose estimation network (denoted as $\mathsf{head}_D^{pose}$). 

We give an overview of how our depth completion is trained, and for more details, we refer readers to the original paper~\cite{ma2019self} that does sparse-to-dense depth completion. We denote the current RGB image as $i_t$ and the nearby RGB image as $i_{t \pm 1}$ (the adjacent RGB image\footnote{the previous or next frame} in video sequences, in which ARKitScenes considers $10$Hz RGB images) and their corresponding sparse Lidar points as ${sl}_{t}$ and ${sl}_{t \pm 1}$. Then, the predicted dense depth $\hat{d}_{t}$ of $i_{t \pm 1}$ is obtained by 
\begin{equation*}
\small 
\hat{d}_{t} =  \mathsf{head}_D^{depth}\Big(\mathsf{backbone}_{\rm RGB}(i_{t}) , \mathsf{backbone}_{\rm sp\_D}( {sl}_{t} )\Big),
\end{equation*}
and the transformation pose $p_{t\rightarrow t\pm 1}$ from current to nearby image is obtained by 
\begin{equation*}
\small
\begin{split}
   p_{t\rightarrow t\pm 1} =  \mathsf{head}_D^{pose}\Big( & \mathsf{backbone}_{\rm RGB}(i_{t}) , \mathsf{backbone}_{\rm sp\_D}( {sl}_{t} ), \\ & \mathsf{backbone}_{\rm RGB}(i_{t\pm 1}) , \mathsf{backbone}_{\rm sp\_D}( {sl}_{t\pm 1} ) \Big).
\end{split}
\end{equation*}

Along with the camera intrinsics matrix, we are able to perform inverse warping~\cite{jain1989fundamentals} to obtain warped images ${{\rm warped}\,i}_{t\pm 1}$ from current to nearby image: 
{\small 
$$
{{\rm warped}\,i}_{t\pm 1} = {\rm inverse\,\,warping}\Big( \hat{d}_{t} , p_{t\rightarrow t\pm 1} \Big). 
$$}

Finally, we consider the depth completion loss to be a combination of sparse depth supervision ($\mathcal{L}_{\rm sparse \,\,depth}$) and photometric loss ($\mathcal{L}_{\rm photo}$):
\begin{equation*}
\small
\begin{split}
    &\mathcal{L}_{\rm Depth\,\,Completion} = \\ & \mathcal{L}_{\rm sparse \,\,depth}(\hat{d}_{t}, {sl}_{t}) + \mathcal{L}_{\rm photo} ({{\rm warped}\,i}_{t\pm 1}, {i}_{t \pm 1}).
\end{split}
\end{equation*}

\subsection{Semantic Segmentation}
\label{subsec:sem_seg}

Since we do not have semantic segmentation labels in ARKitScenes, we consider a knowledge distillation setup using the Mask2Former~\cite{cheng2021masked} semantic segmentation model that is pre-trained on ADE20K dataset~\cite{zhou2019semantic}. In particular, we denote teacher semantic segmented labels ${\rm sem\_seg}^{\rm teacher}_{t}$ for the current RGB image ($i_{t}$) as 
{\small
$${\rm sem\_seg}^{\rm teacher}_{t} = {\rm Mask2Former}(i_{t}).
$$}
Our semantic segmentation network comprises RGB-image backbone ($\mathsf{backbone}_{\rm RGB}$), RGB-image neck (composed of three BiFPN~\cite{tan2020efficientdet} layers and denoted as $\mathsf{neck}_{\rm RGB}$), and a semantic segmentation head (denoted as $\mathsf{head}_{SS}$). In MMISM, following Mask2Former, we consider an additional convolutional neural network as $\mathsf{head}_{SS}$. Then, student semantic segmented labels ${\rm sem\_seg}^{\rm student}_{t}$ are generated as
{\small $$
{\rm sem\_seg}^{\rm student}_{t} = \mathsf{head}_{SS}\Big(\mathsf{neck}_{\rm RGB}\big(\mathsf{backbone}_{\rm RGB}(i_{t} )\big)\Big).
$$}
Lastly, we consider a knowledge distillation loss ($\mathcal{L}_{\rm knowledge}$) in literature~\cite{hinton2015distilling} for the semantic segmentation loss $\mathcal{L}_{\rm Semantic\,\, Segmentation}$:
\begin{equation*}
\begin{split}
& \mathcal{L}_{\rm Semantic\,\, Segmentation} = \\ & \mathcal{L}_{\rm knowledge}({\rm sem\_seg}^{\rm teacher}_{t}, {\rm sem\_seg}^{\rm student}_{t}).    
\end{split}
\end{equation*}

\subsection{3D Object Detection}
\label{subsec:object_detec}

Differing from previous tasks, 3D object detection considers a point cloud backbone and head. As suggested in the ARKitScenes paper~\cite{baruch2021arkitscenes}, we use the VoteNet~\cite{qi2019deep} network to perform 3D object detection. Specifically, we consider the PointNet++~\cite{qi2017pointnet++} network as the feature backbone (denoted as $\mathsf{backbone}_{\rm PC}$) and the VoteNet head as the 3D object detection head (denoted as $\mathsf{head}_{O}$).

For the first step, MMISM takes the predicted dense depth from the depth completion task $\widehat{d}_{t}$, concatenating the corresponding RGB values with the image $i_t$. Then, we perform backward projection using the camera intrinsic and extrinsic matrices on the concatenated RGB-D images to obtain the colored point clouds:
{\small $$
{\rm color\_pc}_t = {\rm backward\,projection} \,([\widehat{d}_{t}, i_t]).
$$}
Next, we obtain point clouds features ${\rm pc\_feat}_t$ by passing ${\rm color\_pc}_t$ into the point clouds backbone $\mathsf{backbone}_{\rm PC}$:
{\small
$$
{\rm pc\_feat}_t = \mathsf{backbone}_{\rm PC} ({\rm color\_pc}_t).
$$
}
Since semantic segmentation and object detection are correlated, a straightforward way to improve object detection is by referencing the output from the semantic segmentation task. Note that our point cloud features ${\rm pc\_feat}_t$ are a sparse set of high-level features of point clouds. To fuse  ${\rm pc\_feat}_t$ and the semantic segmentation predictions ${\rm sem\_seg}^{\rm student}_{t}$, we first query pixel coordinates given the 3D locations of ${\rm pc\_feat}_t$\footnote{Given camera parameters, we can obtain pixel coordinates from 3D locations using forward projection.}, and then concatenate ${\rm sem\_seg}^{\rm student}_{t}$ on those pixel coordinates with ${\rm pc\_feat}_t$. Last, we can obtain the predicted 3D bounding boxes $\widehat{\rm box}_{t}$: 
{\small $$
\widehat{\rm box}_{t} = \mathsf{head}_{O}\Big(\big[{\rm pc\_feat}_t, {\rm queried}({\rm sem\_seg}^{\rm student}_{t})\big])\Big). 
$$}
The object detection loss is then computed using the Hough voting loss $\mathcal{L}_{\rm vote}$~\cite{qi2019deep} between true 3D bounding boxes ${\rm box}_{t}$ and the generated boxes $\widehat{\rm box}_{t}$:
{\small $$
\mathcal{L}_{\rm Object\,\,Detection} = \mathcal{L}_{\rm vote}({\rm box}_{t}, \widehat{\rm box}_{t}).
$$}

\subsection{Human Pose Estimation}
\label{subsec:hum_pose}

Since ARKitScenes does not contain humans, we need to leverage an additional dataset for human pose estimation, for which we select the MS-COCO human pose dataset~\cite{lin2014microsoft}. We use the same RGB-image backbone $\mathsf{backbone}_{\rm RGB}$ and RGB-image neck $\mathsf{neck}_{\rm RGB}$ and consider the CenterNet~\cite{zhou2019objects} human pose estimation head (denoted as $\mathsf{head}_H$). We obtain the estimated human pose $\widehat{\rm pose}_{t}$:
{\small $$
\widehat{\rm pose}_{t} = \mathsf{head}_{H}\Big(\mathsf{neck}_{\rm RGB}\big(\mathsf{backbone}_{\rm RGB}(i_{t} )\big)\Big).
$$}
The human pose estimation loss is computed using the regression loss ($\mathcal{L}_{\rm regression}$) described in the CenterNet paper~\cite{zhou2019objects} between true human poses ${\rm pose}_{t}$ and the generated ones $\widehat{\rm pose}_{t}$:
{\small $$
\mathcal{L}_{\rm Human\,\,Pose\,\,Estimation} = \mathcal{L}_{\rm regression} ({\rm pose}_{t}, \widehat{\rm pose}_{t}).
$$}

\section{Experiments}
\label{sec:exp}

In this section, we show qualitative and quantitative results for our {\bf M}ulti-modality input {\bf M}ulti-task output {\bf I}ndoor {\bf S}cene understanding {\bf M}odel (MMISM). For depth completion, semantic segmentation, and 3D object detection tasks, we perform training and evaluation on ARKitScenes dataset~\cite{baruch2021arkitscenes} using its released training and evaluation split. For the human pose estimation task, we train and evaluate on the MS-COCO human pose estimation dataset~\cite{lin2014microsoft}. We also provide a generated video using MMISM in an indoor office in Supplementary.

\subsection{Quantitative Results}
\label{subsec:quantitative}

\begin{table}[t!]
\vspace{1mm}
\centering
\scalebox{0.9}{
\begin{tabular}{c|c|cc|c}
\toprule
\multirow{2}{*}{Methods} & \multirow{2}{*}{multi-task}  & \multicolumn{2}{c|}{{\bf Error}} & {\bf Accuracy}               \\  
                        & & AbsRel ($\downarrow$)        & RMS ($\downarrow$)            & $\delta < 1.25$ ($\uparrow$)                \\  \midrule \midrule  \multicolumn{5}{c}{\it input: monocular images}  \\ \midrule \midrule 
MonoIndoor~\cite{ji2021monoindoor}  & \xmark          & 0.238           & 0.529          & 0.645\\  \midrule \midrule  \multicolumn{5}{c}{\it input: monocular images + sparse lidar points}  \\ \midrule \midrule 
Sparse-to-Dense~\cite{ma2019self} & \xmark           & 0.048           & 0.115          & 0.977                   \\
MMISM (ours)     & \cmark        &  0.041 &  0.101 &  0.983 \\
\bottomrule
\end{tabular}
}
\caption{Quantitative results for depth completion.}
\vspace{-5mm}
\label{tbl:depth}
\end{table}

{\bf Depth completion task.} Following prior literature~\cite{eigen2014depth}, we evaluate the performance of depth completion using the following metrics: mean absolute relative error (AbsRel), root mean squared error (RMS), and the accuracy under threshold $\delta < {1.25}$. The results are presented in Table~\ref{tbl:depth}. We consider two baseline methods: MonoIndoor~\cite{ji2021monoindoor} and the Self-supervised Sparse-to-Dense~\cite{ma2019self} method. MonoIndoor performs depth estimation based on monocular images only. Self-supervised Sparse-to-Dense performs depth completion from both monocular images and sparse Lidar points. We note that Self-supervised Sparse-to-Dense is a special case of MMISM: when performing a single-task training on depth completion in our model.

First, we compare methods that utilize only monocular images versus monocular images and sparse Lidar points. We observe an obvious performance improvement from MonoIndoor to Self-supervised Sparse-to-Dense on AbsRel ($0.238$ to $0.048$). The improvement suggests 1) indoor depth estimation is a challenging problem and 2) sparse Lidar points can greatly help improve the estimated depth even when these Lidar points are collected from accessible mobile hardware. Next, we compare single-task-training and multi-task-training models. We find AbsRel improves from Self-supervised Sparse-to-Dense to MMISM ($0.048$ to $0.041$), which indicates there are benefits of jointly-training multiple correlated tasks.

\begin{table}[t!]
\centering
\scalebox{1.0}{
\begin{tabular}{c|c|c}
\toprule
Methods & multi-task  &  Acc. ($\uparrow$)  \\   \midrule  
Knowledge Distillation~\cite{hinton2015distilling} & \xmark                   & 0.609    \\
MMISM (ours)     & \cmark        & 0.610 \\
\bottomrule
\end{tabular}
}
\caption{Quantitative results for semantic segmentation.}
\vspace{-6mm}
\label{tbl:semseg}
\end{table}

{\bf Semantic segmentation task.} Recall that, for the semantic segmentation task, we perform knowledge distillation from the Mask2Former~\cite{cheng2021masked} model trained on the ADE20K~\cite{zhou2019semantic} dataset. 
Since we do not have semantic segmentation labels in ARKitScenes, we instead evaluate the knowledge distillation performance. We treat the labels generated from the teacher model (Mask2Former) as true labels and compute the accuracy of the labels generated from the student model (MMISM). We consider the single-task training variant of MMISM as a baseline, reporting numbers in the left of Table~\ref{tbl:semseg}. We find that the multi-task training model performs on par with the single-task training model.

\begin{table}[t!]
\vspace{1mm}
\centering
\scalebox{0.96}{
\begin{tabular}{c|c|cc|c}
\toprule
\multirow{2}{*}{Methods} & \multirow{2}{*}{multi-task}  & RGB & semantic & \multirow{2}{*}{mAP@0.25 IoU ($\uparrow$)}              \\  
                        & & input        & fusion            &    \\  \midrule 
    VoteNet~\cite{qi2019deep} &  \xmark & \xmark & \xmark & 0.426  \\ \midrule
    \multirow{3}{*}{MMISM (ours)} & \multirow{3}{*}{\cmark} & \cmark & \xmark & 0.468 \\ 
      &  & \xmark & \cmark & 0.433              \\
     &  & \cmark & \cmark & 0.476 \\
\bottomrule
\end{tabular}
}
\caption{Quantitative results for 3D object detection.}
\vspace{-4mm}
\label{tbl:object}
\end{table}

\begin{table}[t!]
\centering
\scalebox{1.0}{
\begin{tabular}{c|c|c}
\toprule
Methods  & multi-task  &  mAP@0.75 IoU ($\uparrow$) \\   \midrule  
CenterNet~\cite{zhou2019objects} &  \xmark                   & 0.474     \\
MMISM (ours)   & \cmark        & 0.499 \\
\bottomrule
\end{tabular}
}
\caption{Quantitative results for human pose estimation.}
\vspace{-4mm}
\label{tbl:human}
\end{table}

{\bf 3D object detection task.} We choose the evaluation metric to be the average precision with 3D IoU threshold $0.25$ as proposed by prior literature~\cite{baruch2021arkitscenes,qi2019deep,song2015sun} between the estimated 3D bounding boxes and true 3D bounding boxes. For the baseline method, we consider the original VoteNet~\cite{qi2019deep} method in ARKitScenes~\cite{baruch2021arkitscenes}, representing a model that performs single-task-training, without RGB images in the input, and without fusing the predictions from the semantic segmentation task. We also consider other variants of MMISM by removing the RGB-images from the input or removing the fusion from the semantic segmentation predictions. We present results in Table~\ref{tbl:object}.

First, we observe performance improvements from VoteNet to MMISM's variants, which indicates that multi-task training can improve 3D object detection. Second, we compare different variants of MMISM. We find that removing the RGB-images from the input has the largest affect on performance, which drops from $0.476$ to $0.433$. Nonetheless, the fusion from the semantic segmentation predictions also helps to improve the model, which improves from $0.468$ to $476$ when considering semantic fusion as the changing factor. In a nutshell, both multi-task training and cross-task fusion help improve 3D object detection, with a $11.7\%$ gain from $0.426$ to $0.476$.

{\bf Human pose estimation task.} We follow CenterNet~\cite{zhou2019objects} evaluation protocols to report average precision at IoU thresholds 0.75 between estimated and true human key points. We consider a baseline that performs only the single-task training. Results are presented in Table~\ref{tbl:human}. The results are consistent with the previous tasks, showing that the multi-task training performs on par or better than the single-task training, where the performance improves from $0.474$ to $0.499$.

\begin{figure*}[t!]
    \vspace{1mm}
    \centering
    \includegraphics[width=0.83\linewidth]{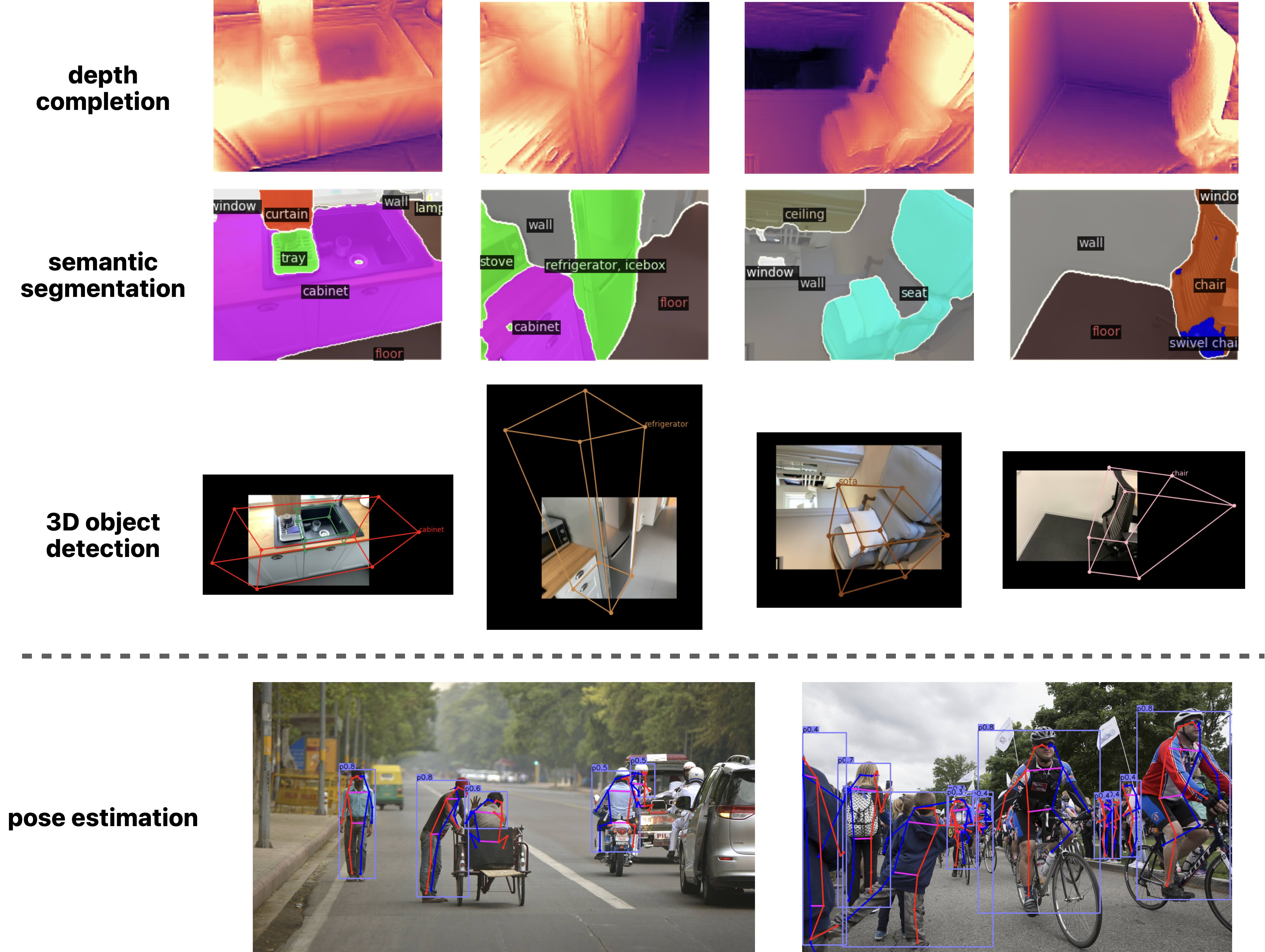}
    \caption{Qualitative results of our model : (a) depth completion, semantic segmentation, and 3D object detection on ARKitScenes dataset~\cite{baruch2021arkitscenes} and (b) human pose estimation on MS-COCO human pose dataset~\cite{lin2014microsoft}. Best viewed in color and zoomed-in.}
    \label{fig:qualitative}
    \vspace{-1mm}
\end{figure*}

\subsection{Qualitative Results}
\label{subsec:qualitative}

In Figure~\ref{fig:qualitative}, we present qualitative results for MMISM. First, we present human pose estimation results on the MS-COCO human pose dataset. Then, we show a sequence of outputs from MMISM on ARKitScenes for depth completion, semantic segmentation, and 3D object detection.

First, we look at the depth completion results. We see that the generated depth map allows us to capture the geometric information in the scene, and we note that we only consider monocular images and sparse Lidar points collected from mobile devices as inputs. Next, we look at the semantic segmentation results. We find there are a few pixels that are wrongly segmented with a jittery output. There are two explanations: 1) the model is applied per frame and there is no temporal module, and 2) the results are generated by performing knowledge distillation from a teacher model trained on a separate dataset. We believe 1) adding a temporal module can improve smoothing of semantic segmentation results, and 2) the segmentation will improve if we have semantic labels for supervised training. Additionally, we find that semantic segmentation predictions assume the ceiling is always on top. This assumption is flawed when the camera is rotated heavily (the third column in Figure~\ref{fig:qualitative}). Then, we look at the 3D object detection results. We see that MMISM is able to perform accurate 3D object detection. Nonetheless, in the Supplementary video, we see the predictions jitter from frame to frame, and we believe that adding a temporal module can alleviate this jittering problem. Last, we look at the human pose estimation results. We find that, even if the environment contains a lot of people, our model can still identify the key points successfully. 

\vspace{-1mm}
\section{Conclusion}
\label{sec:conclu}
\vspace{-1mm}

In this paper, we present {\bf M}ulti-modality input {\bf M}ulti-task output {\bf I}ndoor {\bf S}cene understanding {\bf M}odel (MMISM) for indoor mobile agents. We tackle a critical challenge in indoor scene understanding: insufficient or even no labeled data in real-world indoor environments, by considering a mixture of 1) standard supervised training, 2) self-supervised training, 3) knowledge distillation, and 4) training using external datasets. Our design also addresses other challenges such as fusion between multi-sensory inputs, fusion between multi-task outputs, and computational efficiency. We train, evaluate, and demonstrate the efficacy of MMISM on the ARKitScenes dataset, since its data is collected using accessible mobile hardware. We believe this work sheds light on the advantage of designing multi-modal multi-task systems for indoor scene understanding. We hope our work inspires research and development on intelligent indoor mobile agents, such as fields in AR/VR, robotics, and home security systems.

We consider three future directions. First, we will deploy our inference to on-device mobile computes. Second, we plan to add a temporal module for MMISM to enable more tasks, such as tracking and human motion prediction. 
Third, we will connect MMISM with downstream tasks, such as using MMISM for human robot interaction. 

\newpage

\newpage
{
\small
\bibliography{ref_corl}
\bibliographystyle{plain}
}

\end{document}